\documentclass[conference]{IEEEtran_custom}

\pdfoutput=1 

\usepackage{multicol}
\usepackage{multirow}
\usepackage{comment}
\usepackage{amssymb,amsmath, amsfonts}
\usepackage{array}
\usepackage[hidelinks]{hyperref}
\usepackage{epsfig}
\usepackage{graphicx}
\usepackage[tight]{subfigure}
\usepackage{tabularx}
\usepackage{booktabs}
\usepackage{xspace} 
\usepackage{flushend}
\usepackage{color}
\usepackage[numbers]{natbib}
\usepackage{subfigure}
\usepackage{textcomp}

\usepackage[disable,textsize=scriptsize]{todonotes}

\newcolumntype{M}[1]{>{\arraybackslash}m{#1}}
\newcolumntype{C}[1]{>{\centering \arraybackslash}m{#1}}
\newcolumntype{N}{@{}m{0pt}@{}}

\bibpunct{[}{]}{,}{n}{}{;}

\hypersetup{pdfinfo={
Title={Navigational Instruction Generation as Inverse Reinforcement Learning 
    with Neural Machine Translation},
Author={Andrea F. Daniele, Mohit Bansal, and Matthew R. Walter},
}}

\begin{document}

\title{Navigational Instruction Generation\\ as Inverse Reinforcement Learning\\ with Neural Machine Translation}


\author{
\IEEEauthorblockN{Andrea F. Daniele}
\IEEEauthorblockA{TTI-Chicago, USA\\
{\tt afdaniele@ttic.edu}}
\and
\IEEEauthorblockN{Mohit Bansal}
\IEEEauthorblockA{UNC Chapel Hill, USA\\
{\tt mbansal@cs.unc.edu}}
\and
\IEEEauthorblockN{Matthew R. Walter}
\IEEEauthorblockA{TTI-Chicago, USA\\
{\tt mwalter@ttic.edu}}
}

\maketitle

\begin{abstract}
    Modern robotics applications that involve human-robot interaction
    require robots to be able to communicate with humans seamlessly
    and effectively. Natural language provides a flexible and
    efficient medium through which robots can exchange information
    with their human partners. Significant advancements have been made
    in developing robots capable of interpreting free-form
    instructions, but less attention has been devoted to endowing
    robots with the ability to generate natural language. We propose a
    navigational guide model that enables robots to generate natural
    language instructions that allow humans to navigate a priori
    unknown environments. We first decide which information to share
    with the user according to their preferences, using a policy
    trained from human demonstrations via inverse reinforcement
    learning. We then ``translate'' this information into a natural
    language instruction using a neural sequence-to-sequence model
    that learns to generate free-form instructions from natural
    language corpora. We evaluate our method on a benchmark route
    instruction dataset and achieve a BLEU score of 72.18\% when
    compared to human-generated reference instructions. We
    additionally conduct navigation experiments with human
    participants that demonstrate that our method generates
    instructions that people follow as accurately and easily as those
    produced by humans.
\end{abstract}

\section{Introduction}

\begin{figure}[!t]
  \centering
  \vspace{12pt}
  \begin{tabular}{ M{8.4cm} N}
    \toprule
    \textbf{Input:} map and path & \\[3pt]
    \centering
    \vspace{0.1cm}
  	\includegraphics[width=0.92\linewidth]{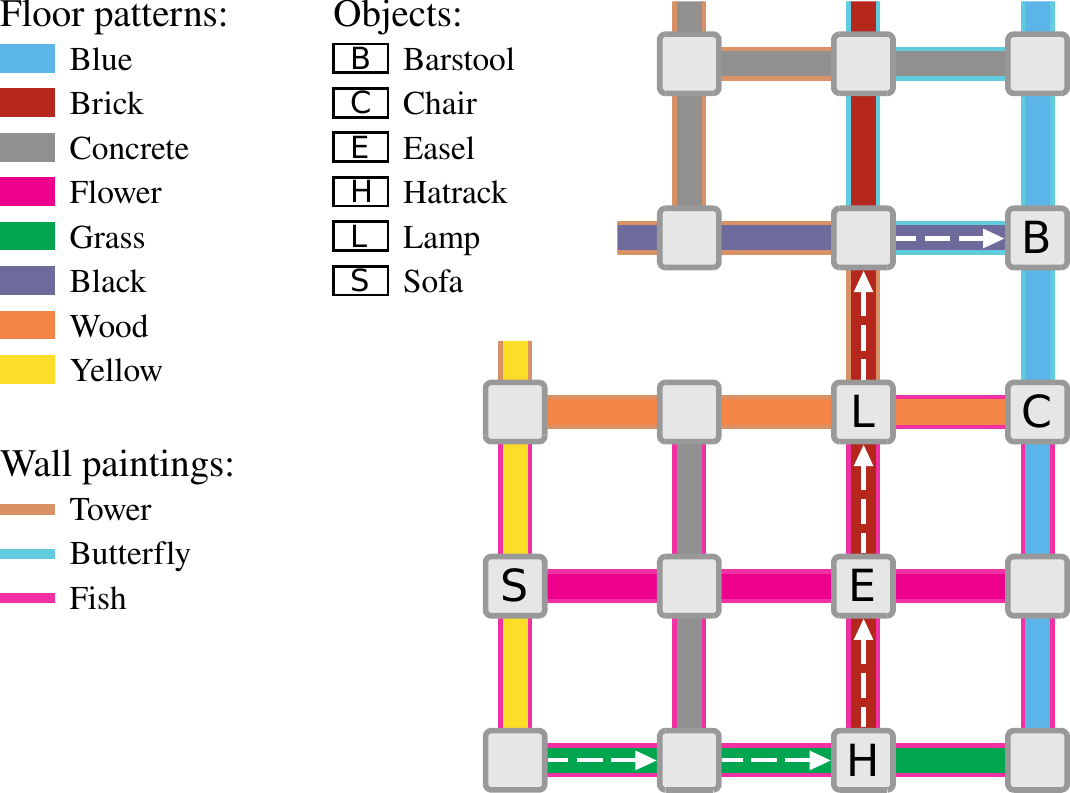}
  	\vspace{0.1cm}
  	& \\
    \midrule
    \textbf{Output:} route instruction & \\[3pt]
    \emph{``turn to face the grass hallway. walk forward twice. face the easel. move until you see black floor to your right. face the stool. move to the stool''}
  	& \\[26pt]
    \bottomrule
  \end{tabular}
  \caption{An example route instruction that our framework
    generates for the shown map and path.\label{fig:generation_example}}
\end{figure}
Robots are increasingly being used as our partners, working with
and alongside people, whether it is serving as assistants
in our homes~\cite{walters07}, transporting cargo in
warehouses~\cite{correa10}, helping students with language learning in the
classroom~\cite{kanda04}, and acting as guides in public
spaces~\cite{hayaski07}. In order for 
humans and robots to work together effectively, robots 
must be able to communicate
with their
human partners in order to establish a shared understanding of the
collaborative task and to coordinate their efforts~\cite{grosz96,
  fong01, clair15, sauppe15}.
Natural language provides an efficient,
flexible medium through which humans and robots can exchange
information. Consider, for example, a search-and-rescue operation
carried out by a human-robot team. The human may first issue spoken
commands (e.g., ``Search the rooms at the end of the hallway'') that
direct one or more robots to navigate throughout the building
searching for occupants~\cite{matuszek10, tellex2011understanding,
  mei2016listen}. In this process, the robot may engage the user in
dialogue to resolve any ambiguity in the task (e.g., to clarify which
hallway the user was referring to)~\cite{tellex12a, deits13, raman13,
  tellex2014asking, hemachandra15a}. The user's ability to trust their
robotic partners is also integral to effective collaboration~\cite{groom07}, and a
robot's ability to generate natural language explanations of its
progress (e.g., ``I have inspected two rooms'') and decision-making
processes have been shown to help establish trust~\cite{dzindolet03,
andrist13, wang16}.

In this paper, we specifically consider the surrogate problem of
synthesizing natural language route instructions and describe a method
that generates free-form directions that people can accurately and
efficiently follow in environments unknown to them a priori 
(Fig.~\ref{fig:generation_example}). This specific problem has previously
been considered by the robotics community~\cite{goeddel2012dart,
oswald2014learning} and is important for human-robot
collaborative tasks, such as search-and-rescue, exploration, and
surveillance~\cite{kunze14}, and for robotic assistants, such as those
that serve as guides in museums, offices, and other public
spaces. More generally, the problem is relevant beyond human-robot
interaction to the broader domain of indoor navigation, for which
GPS is
unavailable and the few existing solutions that rely upon template-based 
instructions, referring to distances and street names, are not suitable. 
There are two primary challenges to generating effective, natural 
language route instructions, which are characteristic of the more general 
problem of free-form generation. 

The first challenge is \emph{content selection}, the problem of deciding what 
and how much information to convey to the user as part of the
directions. In general, the more detailed an instruction is, the less
ambiguous it is. However, verbose instructions can be unnatural and hard for
followers to remember and, thus, ineffective. Consequently, it is
important to balance the value of including particular information as
part of a route instruction with the cost that comes with
increasing the level of detail. Further, not all information is
equally informative. Existing commercial navigational solutions typically rely 
on a set of hand-crafted rules that consider only street names and
metric distances as valid candidates, the latter of which requires
that follower's keep track of their progress. In contrast, studies have shown that
people prefer route instructions that reference physical, salient
landmarks in the environment~\cite{waller2007landmarks}. However, no
standard exists with regards to what and how these landmarks should be
selected, as these depend on the nature of the environment and the
demographics of the follower~\cite{ward86, hund2012impact}. 

We propose a method that models this content selection problem as a
Markov decision process with a learned policy that decides what and
how much to include in a formal language specification of the task
(path). We learn this policy via inverse reinforcement learning from
demonstrations of route instructions provided by humans. This avoids
the need for hand-crafted selection rules and allows our method to 
automatically adapt to the preferences and communication style of the target
populations and to simultaneously choose to convey information that
minimizes the ambiguity of the instruction while avoiding verbosity.

The second challenge is \emph{surface realization}, which is
the task of synthesizing a natural language sentence that refers to
the content selected in the first step. Existing solutions rely on sentence templates,
generating sentences by populating manually defined fields (e.g.,
``turn $\langle$direction$\rangle$'') and then serializing these
sentences in a turn-by-turn fashion. As expected, the use of such templates reduces
coherence across sentences and limits the ability to adapt to
different domains (e.g., from outdoor to indoor navigation).  Additionally, while
the output is technically correct, the resulting sentences tend to be
rigid and unnatural.  Studies show that language generated by a robot
is most effective when it emulates the communication style that
people use~\cite{torrey13}. 

We address the surface realization problem through a neural
sequence-to-sequence model that ``translates'' a formal language
specification of the selected command into a natural language
sentence. Our model takes the form of an encoder-aligner-decoder
architecture that first encodes the formal path specification with a
recurrent neural network using long short-term memory
(LSTM-RNN)~\cite{hochreiter97}. The model then decodes (translates) the 
resulting abstraction of the input into a natural language sentence (word
sequence), using an alignment mechanism to further refine the selected
information and associate output words with the corresponding elements
in the input formal specification. The use of LSTMs as the hidden units
enables our model to capture the long-term dependencies that exist
among the selected information and among the words in the resulting
instruction. We train our surface realization model on
instruction corpora, which enables our method to generate natural
language directions that emulate the style of human instructions, without 
the need for templates, specialized features, or linguistic resources.

We evaluate our method on the benchmark SAIL dataset of human-generated route
instructions~\cite{macmahon2006walk}. Instructions generated with our
method achieve a sentence-level BLEU score of $72.18\%$, indicating their similarity
with the reference set of human-provided instructions. We perform
a series of ablations and visualizations to better understand the
contributions of the primary components of our model. We additionally
conduct human evaluation experiments and demonstrate that our learning-based
method generates instructions that people are able to follow as
efficiently and accurately as those generated by humans. A qualitative
assessment reveals that participants rate the quality of information
conveyed by our instructions, the ease with which they are
interpretable, and the participants' confidence in following the
instructions equivalent to and even better than those of human-generated directions.
\section{Related Work}
\label{sec:related_work}

Existing research related to the generation of route instructions
spans the fields of robotics, natural language processing, cognitive
science, and psychology. Early work in this area focuses on
understanding the way in which humans generate natural language route
instructions~\cite{ward86, allen97, lovelace1999elements} and the properties
that make ``good'' instructions easier for people to
follow~\cite{look2005location, richter2008simplest, waller2007landmarks}. 
These studies have shown that people prefer to
give directions as a sequence of turn-by-turn instructions and that
they favor physical objects and locations as intuitive landmarks
(as opposed to the quantified distances typical of GPS-based
navigation systems).

Based on these studies, much of the existing research on
generating route instructions involves the use of hand-crafted rules
that are designed to emulate the manner in which people compose
navigation instructions~\cite{striegnitz11, curry15}. \citet{look2005location} 
compose route instructions using a set of templates and application rules 
engineered based upon a corpus of human-generated route instructions.
\citet{look2008cognitively} improves upon this work by incorporating
human cognitive spatial models to generate high-level route overviews
that augment turn-by-turn directions. Similarly,~\citet{dale2005using}
analyze a dataset of route instructions composed by people to derive a
set of hand-designed rules that mimic the content and style of human
directions. \citet{goeddel2012dart} describe a particle filter-based
method that employs a generative model of direction following to
produce templated  instructions that maximize the likelihood of
reaching the desired destination.

The challenge with instruction generation systems that rely upon
hand-crafted rules is that is difficult to design a policy that
generalizes to a wide variety of scenarios and followers, whose
preferences vary depending on such factors as their cultural
background~\cite{hund2012impact} and
gender~\cite{ward86}. \citet{cuayahuitl10} seek to improve upon this
using reinforcement learning with hand-crafted reward functions that
model the length of the instructions and the likelihood that they will
confuse a follower. They then learn a policy that reasons both over
the best route and the corresponding navigational
instructions. However, this approach still requires that domain
experts define the reward functions and specify model parameters. In
contrast,~\citet{oswald2014learning} model the problem of deciding
what to include in the instruction (i.e., the content selection
problem) as a Markov decision process and learn a policy from a
human-written navigation corpus using maximum entropy inverse
reinforcement learning. Given the content identified by the policy,
their framework does not perform surface realization, and instead
generates instructions by matching the selected content with the
nearest match in a database of human-generated instructions. Our
method also uses inverse reinforcement learning for content
selection, but unlike their system, our method also learns to perform
surface realization directly from corpora, thus generating newly-composed 
natural language instructions.

Relatedly, much attention has been paid recently to the ``inverse''
problem of learning to follow (i.e., execute) natural language route
instructions. Statistical methods primarily formulate the problem of
converting instructions to actions as either a semantic parsing 
task~\cite{matuszek10,chen11,artzi2013weakly} or as a symbol grounding
problem~\cite{kollar2010toward, tellex2011understanding, landsiedel13,
howard14a, chung15}. Alternatively, \citet{mei2016listen} learn to translate
natural language instructions to action sequences via an end-to-end fashion 
using an encoder-aligner-decoder architecture.

Meanwhile, selective generation considers the more general problem of
converting a rich database to a natural language utterance, with
existing methods generally focusing on the individual problems of
content selection and surface realization.  \citet{barzilay2004catching}
perform content selection on collections of unannotated documents for
the sake of text summarization. \citet{barzilay2005collective}
formulate content selection as a collective classification problem,
simultaneously optimizing local label assignments and their pairwise
relations. \citet{liang2009learning} consider the related problem of
aligning elements of a database to textual description clauses. They
propose a generative semi-Markov model that simultaneously segments
text into utterances and aligns each utterance with its corresponding
entry in the database. Meanwhile, \citet{walker01} perform surface
realization via sentence planners that can be trained to generate
sentences for dialogue and context
planning. \citet{wong2007generation} effectively invert a semantic
parser to generate natural language sentences from formal meaning
representations using synchronous context-free grammars. Rather than
consider individual sub-problems, recent work focuses on solving
selective generation via a single framework~\cite{chen2008learning,
  kim2010generative, angeli2010simple, konstas2012unsupervised,
  mei2016what}. \citet{angeli2010simple} model  content selection
and surface realization as local decision problems via log-linear
models and employ templates for generation. \citet{mei2016what}
formulate selective generation as an end-to-end learning problem and
propose a recurrent neural network encoder-aligner-decoder model that
jointly learns to perform content selection and surface realization
from database-text pairs.
\section{Task Definition}
\label{sec:task_definition}

We consider the problem of generating natural language instructions
that allow humans to navigate environments that are unknown to them a
priori. As with the broader class of language generation problems,
this task requires deciding which information to convey to the user
(content selection), such that it is correct (e.g., consistent with
what is currently visible to the user in the environment), not overly
verbose (i.e., so that users can easily interpret and remember the
instruction), and unambiguous. The task then requires
conveying this information via language, such
that the sentence is syntactically correct, its semantics are
consistent with the intended message, and it is natural and free-form.

Formally, given a map of the environment and a desired path, the task
is to produce a natural language instruction that guides
the user along the path. The map $m$ takes the form of a hybrid
metric-topologic-semantic representation
(Fig.~\ref{fig:generation_example}) that encodes the position of and
connectivity between a dense set of locations in the environment
(e.g., intersections) and the position and type of objects and
environment features (e.g., floor patterns). The path $p$ is a sequences 
of poses (i.e., position and orientation) that
corresponds to the minimum distance route from a
given initial pose to a desired goal pose. We split the path
according to changes in direction, representing the path
$p = (p_1, p_2, \ldots p_M)$ as a sequence of intermediate segments
$p_i$.

Training data comes in the form of tuples $(m^{(i)}, p^{(i)},
\Lambda^{(i)})$ for $i=1, 2, \ldots, n$ drawn from human
demonstrations, where $m^{(i)}$ is a map of the environment,
$\Lambda^{(i)}$ is a human-generated natural language route
instruction, and $p^{(i)}$ is the path that a different human took
when following the instructions.
At test time, we consider only the
map and path pair as known and hold out the human-generated
instruction for evaluation. The dataset that we use for training,
validation, and testing comes from the benchmark SAIL
corpus~\cite{macmahon2006walk}.
\section{Model} \label{sec:model}

\begin{figure}[t]
  \centering
  \includegraphics[width=\linewidth]{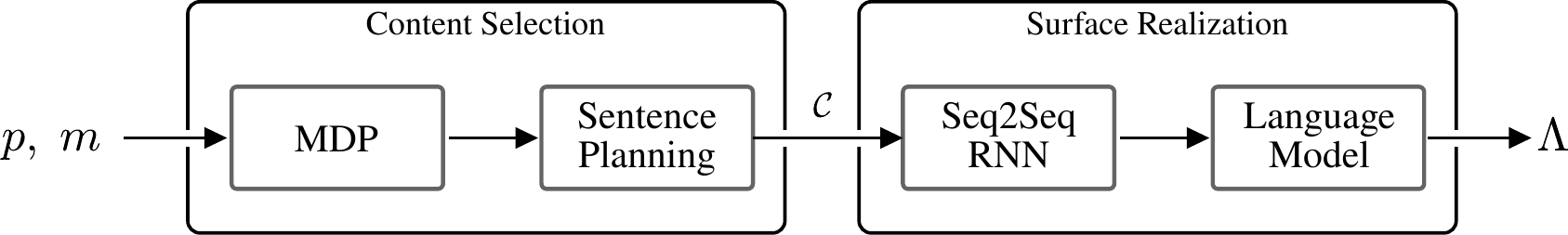}
  \caption{Our method generates natural language instructions for a
    given map and path.} \label{fig:framework}
\end{figure}

Given a map and path, our framework (Fig.~\ref{fig:framework})
performs content selection to decide what
information to share with the human follower and subsequently performs
surface realization to generate a natural language instruction
according to this selected content. Our method learns to perform
content selection and surface realization from human demonstrations,
so as to produce instructions that are similar to those generated by
humans.

\subsection{Compound Action Specifications} \label{sec:cas}

In order to bridge the gap between the low-level nature of the input
paths and the natural language output, we encode paths using an
intermediate logic-based formal language. Specifically, we use the
Compound Action Specification (CAS)
representation~\cite{macmahon2006walk}, which provides a formal
abstraction of navigation commands for hybrid
metric-topologic-semantic maps such as ours. The CAS language consists
of five \textit{actions} (i.e., Travel, Turn, Face, Verify, and
Find), each of which is associated with a number of
\textit{attributes} that together define specific commands (e.g.,
Travel.distance, Turn.direction). We
distinguish between CAS \emph{structures}, which are instructions with
the attributes left empty (e.g.,
\textit{Turn(direction=\textbf{None})}) thereby defining a class of
instructions, and CAS \emph{commands}, which correspond to
instantiated instructions with the attributes set to particular values
(e.g., \textit{Turn(direction=\textbf{Left})}).
For each English instruction $\Lambda^{(i)})$ in the dataset, we generate
the corresponding CAS command $c^{(i)}$ using the MARCO architecture 
\cite{macmahon2006walk}.For a complete
description of the CAS language, see~\citet{macmahon2006walk}.

\subsection{Content Selection} \label{sec:content_selection}

There are many ways in which one can compose a CAS specification of the
desired path, both in terms of the type of information that is conveyed 
(e.g., referencing
distances vs.\ physical landmarks), as well as the specific references
to use (e.g., different objects provide candidate landmarks). Humans
exhibit common preferences in terms of the type of information that is
shared (e.g., favoring visible landmarks over
distances)~\cite{waller2007landmarks}, yet the specific nature of this
information depends upon the environment and the followers'
demographics~\cite{ward86, hund2012impact}. Our goal is to learn these
preferences from a dataset of instructions generated by humans.

\subsubsection{MDP with Inverse Reinforcement Learning} \label{sec:irl}

In similar fashion to \citet{oswald2014learning}, we formulate the content 
selection problem as a Markov decision
process (MDP) with a goal of then identifying an information selection
policy that maximizes long-term cumulative reward consistent with
human preferences (Fig.~\ref{fig:framework}). However, this reward
function is unknown a priori and generally difficult to define. We
assume that humans optimize a common reward function when composing
instructions and employ inverse reinforcement learning to learn a
policy that mimics the preferences that humans exhibit based upon a
set of human demonstrations.

An MDP is defined by the tuple $(S, A, R, P, \gamma)$, where $S$ is a
set of states, $A$ is a set of actions,
$R(s, a, s^\prime) \in \mathbb{R}$ is the reward received when
executing action $a \in A$ in state $s \in S$ and transitioning to
state $s^\prime \in S$, $P(s^\prime \vert a,s)$ is the probability of
transitioning from state $s$ to state $s^\prime$ when executing action
$a$, and $\gamma \in (0,1]$ is the discount factor. The policy
$\pi(a \vert s)$ corresponds to a distribution over actions given the
current state.  In the case of the route instruction domain, the state
$s$ defines the user's pose and path in the context of the map of the
environment. We represent the state in terms of $14$ \emph{context}
features that express characteristics such as changes in orientation
and position, the relative location of objects, and nearby environment
features (e.g., floor color). We encode the state $s$ as a
$14$-dimensional binary vector that indicates which context features
are active for that state. In this way, the state space $S$ is that spanned 
by all possible instantiations of context features. Meanwhile, the action
space corresponds to the space of different CAS structures (i.e.,
without instantiated attributes)  that can be used to define the path.

We seek a policy $\pi(a \vert s)$ that maximizes expected cumulative
reward. However, the reward function that defines the value of
particular characteristics of the instruction is unknown and difficult
to define. For that reason, we frame the task as an inverse
reinforcement learning (IRL) problem using human-provided route
instructions as demonstrations of the optimal policy. Specifically, we learn a
policy using the maximum entropy formulation of IRL~\cite{ziebart08},
which models user actions as a distribution over paths parameterized
as a log-linear model $P(a; \theta) \propto e^{-\theta^\top \xi(a)}$,
where $\xi(a)$ is a feature vector defined over actions. We consider $9$
instruction features (\emph{properties})  that
include features expressing the number of landmarks included in the
instruction, the frame of reference that is used, and the complexity
of the command. The feature vector $\xi(a)$ then takes the form of a
$9$-dimensional binary vector. Appendix~\ref{app:mdp_feature} presents the
full set of context and property features used to parameterize the
state and action, respectively. Maximum entropy IRL then solves for
the distribution via the following optimization
\begin{equation}
    \begin{split}
        &P(a; \theta^*) = \underset{\theta}{\text{arg max}} \; P(a; \theta) \log P(a; \theta)\\
        &\text{s.t.} \; \xi_g = \mathbb{E}[\xi(a)],
    \end{split}
\end{equation}
where $\xi_g$ denotes the features from the demonstrations and the
expectation is taken over the action distribution. For further details
regarding maximum entropy IRL, we refer the reader to \citet{ziebart08}.

\begin{figure*}[t]
    \centering
    \includegraphics[width=0.90\linewidth]{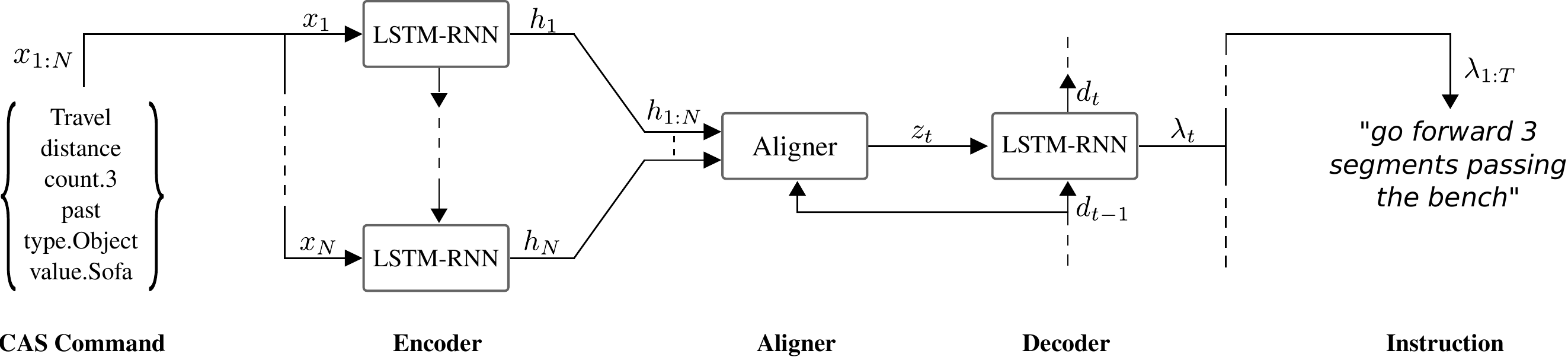}
    \caption{Our encoder-aligner-decoder model for surface realization.
    \label{fig:seq2seq_model}}
\end{figure*}

The policy defines a distribution over CAS structure compositions (i.e., using the
\emph{Verify} action vs.\ the \emph{Turn} action) in terms of their feature encoding. 
We perform inference over this policy to identify the maximum a
posteriori property vector $\xi(a^*) = \text{arg max}_{\xi} \,
\pi$. As there is no way to invert the feature mapping, we then match
this vector $\xi(a^*)$ to
a database of CAS structures formed from our
training set. Rather than choosing the nearest match, which may result
in an inconsistent CAS structure, we retrieve the $k_c$ nearest
neighbors from the database using a weighted distance in terms of
mutual information~\cite{oswald2014learning} that expresses the
importance of different CAS features based upon the context. As
several of these may be valid, we employ spectral clustering using the
similarity of the CAS strings to
identify a set of candidate CAS structures $\mathcal{C}_s$.

\subsubsection{Sentence Planning} \label{sec:sentence_planning}

Given the set of candidate CAS structures $\mathcal{C}_s$, our method next
chooses the attributes
values such that the final CAS commands are both valid and not
ambiguous. We can compute the
likelihood of a command $c$ to be a valid instruction for a path
$p$ defined on a map $m$ as:
\begin{equation} \label{eq:probability_cas}
    P( c \vert p, m ) = 
    \frac{ \delta( c \vert  p, m ) }{ \sum_{j=1}^{K} \delta( c \vert \hat{p}_j, m ) }.
\end{equation}
The index $j$ iterates over all the possible paths that have the same
starting pose of $p$ and $\delta( c \; | \; p, m )$ is defined as:
\begin{equation*}
    \delta( c \vert p, m ) = 
    \left\{
        \begin{array}{ll}
          1 & \text{if} \;\; \eta( c ) = \phi( c, p, m ) \\
          0 & \text{otherwise}
        \end{array}
\right.
\end{equation*}
where $\eta( c )$ is the number of attributes defined in $c$, and
$\phi( c, p, m )$ is the number of attributes defined in $c$ that are
also valid with respect to the inputs $p,m$.

For each candidate CAS structure $c \in \mathcal{C}_s$, we generate multiple 
CAS commands by iterating over the possible attributes values. We evaluate the
correctness and ambiguity of each configuration according to
Equation~\ref{eq:probability_cas}. A command is deemed valid if its
likelihood is greater than a threshold $P_t$. Since the number of possible 
configurations for a
structure increases exponentially with respect to the number of
attributes, we assign attributes using greedy search. The iteration algorithm
is constrained to use only objects and properties of the environment
visible to the follower. The result is a set $\mathcal{C}$ of valid CAS commands.

\subsection{Surface Realization} \label{sec:surface_realization}

Having identified a set of CAS commands suitable to the given path,
our method then proceeds to generate the corresponding natural
language route instruction. We formulate this problem as one of
``translating'' the instruction specification in the formal CAS
language into its natural language equivalent.\footnote{Related
  work~\cite{matuszek10,artzi2013weakly,mei2016listen} similarly models the inverse
  task of language understanding as a machine translation problem.} We
perform this translation using an encoder-aligner-decoder model
(Fig.~\ref{fig:seq2seq_model}) that enables our framework to generate
natural language instructions by learning from examples of
human-generated instructions, without the need for specialized
features, resources, or templates.

\subsubsection{Sequence-to-Sequence Model} \label{sec:seq2seq}

We formulate the problem of generating natural language route
instructions as inference over a probabilistic model $P(\lambda_{1:T}
\vert x_{1:N})$, where $\lambda_{1:T} = (\lambda_1, \lambda_2, \ldots,
\lambda_T)$ is the sequence of words in the instruction and
$x_{1:N} = (x_1, x_2, \ldots x_N)$ is the sequence of tokens in the
CAS command. The CAS sequence includes a token for each action (e.g.,
$Turn$, $Travel$) and a set of tokens with the form \emph{attribute.value} for 
each couple (\textit{attribute,value}); for example, 
\emph{Turn(direction=Right)} is represented by the sequence 
(\emph{Turn}, \emph{direction.Right}). Generating an instruction sequence then corresponds to
inference over this model
\begin{subequations} \label{eqn:argmax}
\begin{align}
    \lambda_{1:T}^* &= 
    	\underset{\lambda_{1:T}}{\textrm{arg max }} P(\lambda_{1:T}|x_{1:N})\\
    &= 
    	\underset{\lambda_{1:T}}{\textrm{arg max }} \prod\limits_{t=1}^T
    	P(\lambda_t \vert \lambda_{1:t-1}, x_{1:N})
\end{align}
\end{subequations}

We model this task as a sequence-to-sequence learning problem, whereby
we use a recurrent neural network (RNN) to first encode the input
CAS command 
\begin{subequations}
   \begin{align}
     h_j &= f(x_j, h_{j-1})\\
     z_t &= b(h_1, h_2, \ldots h_N),
   \end{align}
\end{subequations}
where $h_j$ is the encoder hidden state for CAS token $j$, and $f$ and $b$ are
nonlinear functions, which we define later.  An aligner computes the context vector $z_t$ that
encodes the language instruction at time $t \in \{1,\ldots,T\}$.
An RNN decodes the context vector $z_t$ to arrive at the desired
likelihood (Eqn.~\ref{eqn:argmax})
\begin{equation} \label{eqn:cond_prob}
      P(\lambda_t \vert \lambda_{1:t-1}, x_{1:N}) = g(d_{t-1}, z_t),
\end{equation}
where $d_{t-1}$ is the decoder hidden state at time $t-1$, and $g$ is a nonlinear
function. 

\textbf{Encoder}\hspace{5pt}   Our encoder (Fig.~\ref{fig:seq2seq_model}) takes as input the sequence
of tokens in the CAS command $x_{1:N}$. We transform each token $x_i$ into
a $k_e-$dimensional binary vector using a word embedding representation 
\cite{mikolov2013distributed}. We feed this sequence into an RNN
encoder that employs LSTMs as the recurrent unit as a result of their
ability to learn long-term dependencies among the instruction
sequences, without being prone to vanishing or exploding
gradients. The LSTM-RNN encoder summarizes the relationship between
elements of the CAS command and yields a sequence of hidden states
$h_{1:N} = (h_1, h_2, \ldots, h_N)$, where $h_j$ encodes CAS words up
to and including $x_j$. In practice, we reverse the input sequence
before feeding it into the neural encoder, which has been demonstrated
to improve performance for other neural translation tasks 
\cite{sutskever14}.

Our encoder  is similar to that of \citet{graves13},
\begin{subequations} \label{eqn:encoder}
    \begin{align}
      \begin{pmatrix}
          i^{e}_j\\ 
          f^{e}_j\\
          o^{e}_j\\ 
          g^{e}_j 
      \end{pmatrix} 
   &= 
     \begin{pmatrix}
         \sigma\\
         \sigma\\
         \sigma\\ 
         \tanh 
     \end{pmatrix}
      T^{e} 
      \begin{pmatrix}
          x_j\\ 
          h_{j-1}
      \end{pmatrix}\\
      c^{e}_j&=f^{e}_j \odot c^{e}_{j-1} + i^{e}_j \odot g^{e}_j \\
      h_j &=o^{e}_j \odot \tanh(c^{e}_j) \label{eqn:encoder_h}
    \end{align}
\end{subequations}
where $T^{e}$ is an affine transformation, $\sigma$ is the logistic sigmoid
that restricts its input to $[0,1]$, $i^{e}_j$, $f^{e}_j$, and $o^{e}_j$
are the input, output, and forget gates of the LSTM, respectively, and
$c^e_j$ is the memory cell activation vector. The memory cell $c^e_j$
summarizes the LSTM's previous memory $c^e_{j-1}$ and the current input,
which are modulated by the forget and input gates, respectively. \\

\textbf{Aligner}\hspace{5pt}  Having encoded the input CAS command into a sequence
of hidden annotations $h_{1:N}$, the decoder then seeks to generate a
natural language instruction as a sequence of words.
We employ an alignment mechanism~\cite{bahdanau14} (Fig.~\ref{fig:seq2seq_model}) that
permits our model to match and focus on particular elements of the CAS sequence
that are salient to the current word in the output
instruction. We compute the context vector as
\begin{equation} \label{eqn:attention_model}
    z_t = \sum_{j}\alpha_{tj} h_j.
\end{equation}
The weight $\alpha_{tj}$ associated with the j-th hidden state is
\begin{equation}
    \alpha_{tj} = \exp(\beta_{tj})/\sum_k \exp(\beta_{tk}),
\end{equation}
where the alignment term $\beta_{tk}=f(d_{t-1},h_j)$ expresses the
degree to which the CAS element at position $j$ and those around it
match the output at time $t$. The term $d_{t-1}$ represents the decoder 
hidden state at the previous 
time step. The alignment is modeled as a one-layer neural perceptron
\begin{equation}
  \beta_{tk} = v^\top\tanh(Wd_{t-1}+Vh_{j}),
\end{equation}
where $v$, $W$, and $V$ are learned parameters.\\

\textbf{Decoder}\hspace{5pt}  Our model employs an LSTM decoder
(Fig.~\ref{fig:seq2seq_model}) that takes as input the 
context vector $z_t$ and the decoder hidden state at the previous time step 
$d_{t-1}$ and outputs the conditional probability distribution 
\mbox{$P_{\lambda,t} = P(\lambda_t \vert \lambda_{1:t-1}, x_{1:N})$}
over the next token as a deep output layer
\begin{subequations}
    \begin{align}
      \begin{pmatrix}
          i^d_t\\
          f^d_t\\
          o^d_t\\
          g^d_t
      \end{pmatrix}
   &= 
     \begin{pmatrix}
         \sigma\\ 
         \sigma\\ 
         \sigma\\ 
         \tanh 
     \end{pmatrix} T^d 
      \begin{pmatrix}
          d_{t-1}\\ 
          z_t
      \end{pmatrix}\\
      c^d_t&=f^d_t \odot c^d_{t-1} + i^d_t \odot g^d_t \\
      d_{t}&=o_t^d \odot \tanh(c^d_t)\\
      q_t &= L_0 ( L_dd_t + L_zz_t )\\
      P_{\lambda,t} &= \textrm{softmax}\left(q_t\right)
    \end{align}
\end{subequations}
where $L_0$, $L_d$, and $L_z$ are parameters to be learned.\\

\textbf{Training}\hspace{5pt}  We train our encoder-aligner-decoder 
model so as to predict the natural language instruction $\lambda_{1:T}^*$ for a given 
input sequence $x_{1:N}$ using a training set of human-generated
reference instructions. We use the negative
log-likelihood of the reference instructions at each time step $t$ as
our loss function.
\\

\textbf{Inference}  Given a CAS command represented as a sequence of
tokens $x_{1:N}$, we generate a route instruction as the sequence of
maximum a posteriori words $\lambda_{1:T}^*$  under our learned
model (Eqn.~\ref{eqn:argmax}). We use beam search to perform approximate
inference, but have empirically found greedy search to often perform
better.\footnote{This phenomenon has been observed by
  others~\cite{angeli2010simple, mei2016what}, and we attribute it to
  training the model in a greedy fashion.} For that reason, we generate
candidates using both greedy and beam search.

\subsubsection{Language Model} \label{sec:language_model}

The inference procedure results in multiple candidate instructions for
a given segment, and additional candidates may exist when there are
multiple CAS specifications. We rank these candidate instructions
using a language model (LM) trained on large amounts of English data. We formulate this LM as an LSTM-RNN~\cite{sundermeyer2012lstm} that assigns a perplexity score
to each of the corresponding instructions. 

Given the CAS specifications for a segmented path $p = (p_1, p_2,
\ldots p_M)$, we generate the final instruction $\Lambda$  by
sequencing the $M$ sentences 
$\{\Lambda_1^\star,\ldots,\Lambda_M^\star\}$ (i.e., one for each path segment)
\begin{equation}
    \Lambda_i^\star = \underset{\Lambda_{ij}}{\textrm{arg min }} L(\Lambda_{ij}),
\end{equation}
where $\Lambda_{ij}$ is the j-th candidate for the i-th segment and $L(\Lambda_{ij})$
is the perplexity score assigned by the language model to the sentence $\Lambda_{ij}$.

\section{Experimental Setup} \label{sec:experiment_setup}

\subsection{Dataset} \label{sec:dataset}

We train and evaluate our system using the publicly available SAIL
route instruction dataset collected by~\citet{macmahon2006walk}. We
use the original data without correcting typos or wrong
instructions (e.g., confusing ``left'' and ``right''). The dataset 
consists of $3213$ demonstrations arranged in
$706$ paragraphs produced by 6 instructors for $126$ different paths
throughout $3$ virtual environments, where each demonstration provides
a map-path-command tuple $(m^{(i)}, p^{(i)},
\Lambda^{(i)})$.
We partitioned the dataset into separate training ($70\%$), validation 
($10\%$), and test ($20\%$) sets. 
We use command-instruction pairs $(c^{(i)},
\Lambda^{(i)})$ from the training, 
validation and test sets respectively for training, hyper-parameter 
tuning, and testing of our encoder-aligner-decoder model. 
We use path-command pairs $(p^{(i)},
c^{(i)})$ from the training set for IRL, and pairs from the
validation set to tune the hyper-parameters of the content selection 
model. Finally, we use path-instruction pairs $(p^{(i)},
\Lambda^{(i)})$ from the test 
set to evaluate the performance of our framework through experiments
with human instruction followers.

\subsubsection{Data Augmentation}

The SAIL dataset is significantly smaller than those typically used to
train neural sequence-to-sequence models. In order to overcome this
scarcity, we augmented the original dataset using a set of rules.
In particular, for each command-instruction $(c^{(i)},
\Lambda^{(i)})$ pair in the original dataset
we generate a number of new demonstrations iterating over the set of
possible values for each attribute in the command and updating
the relative instruction accordingly.  For example, given the
original pair (\textit{Turn(direction=Left)},~\textit{``turn left''}), we
augment the dataset with $2$ new pairs, namely
(\textit{Turn(direction=Right)},~\textit{``turn right''}) and
(\textit{Turn(direction=Back)},~\textit{``turn back''}).  Our augmented
dataset consists of about $750$k and $190$k demonstrations for training and
validation, respectively.

\subsection{Implementation Details}
\label{sec:implementation}

We implemented and tested the proposed model using the following
values for the system parameters: $k_c = 100$, $P_t = 0.99$,
$k_e = 128$, and $L_t=95.0$. The encoder-aligner-decoder
consisted of $2$ layers for the encoder and decoder with $128$ LSTM
units per layer. The language model similarly included a $2$-layer
recurrent neural network with $128$ LSTM units per layer. The size of
the CAS and natural (English) language vocabularies was $88$ and
$435$, respectively, based upon the SAIL dataset. All
parameters were chosen based on the performance on 
the validation set.
We train our model using Adam~\cite{kingma2014adam} for
optimization. At test time, we perform approximate inference using a
beam width of two. Our method requires an average of $33$\,s
($16$\,s without beam search) to generate
instructions for a path consisting of $9$ movements when run on a
laptop with a $2.0$\,GHz CPU and $8$\,GB of RAM. As with other neural
models, performance would improve significantly using a GPU.

\subsection{Automatic Evaluation}

To the best of our knowledge, we are the first to use the SAIL dataset
for the purposes of generating route instructions. Consequently, we evaluate 
our method by comparing our generated instructions with a
reference set of human-generated commands from the SAIL dataset using
the BLEU score (a $4$-gram matching-based
precision)~\citep{papineni2002bleu}.  For this purpose, for each
command-instruction pair
($c^{(i)},\Lambda^{(i)}$) in
the validation set, we first feed the command $c^{(i)},$ into our model to
obtain the generated instruction $\Lambda^*$, and secondly use
$\Lambda^{(i)},$ and $\Lambda^*$ respectively as the reference and
hypothesis for computing the 4-gram BLEU score. We consider both the
average of the BLEU scores at the individual sentence level (macro-average
precision) as well as at the full-corpus level (micro-average precision).

\subsection{Human Evaluation} \label{sec:study}
\begin{figure}[!t]
    \centering
    \includegraphics[width=\linewidth]{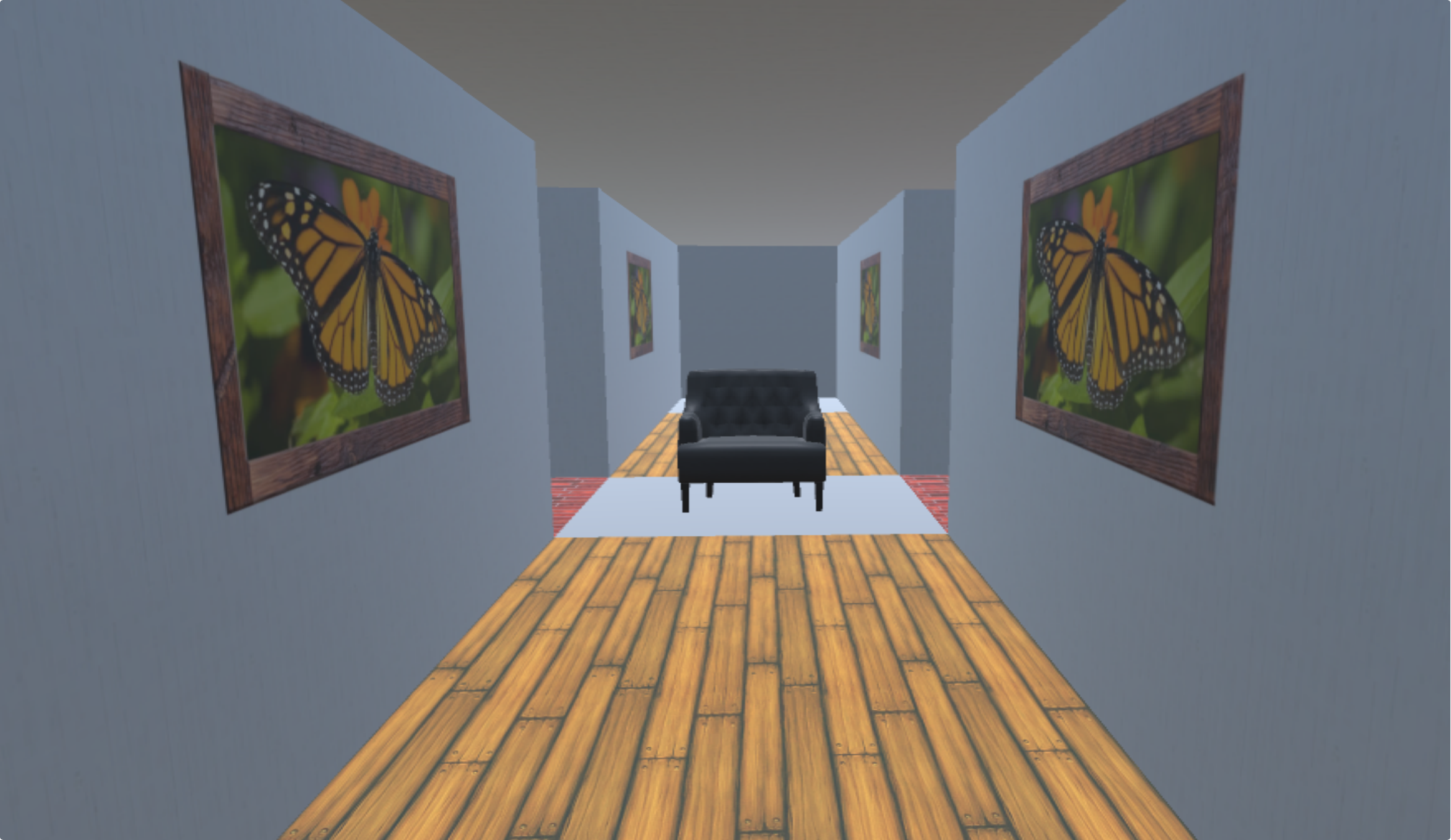}
    \caption{Participants' field of view in the 
      virtual world used for the human navigation experiments.
    \label{fig:simulator}}
\end{figure}
The use of BLEU score indicates the similarity between instructions
generated via our method and those produced by humans, but it does not
provide a complete measure of the quality of the instructions (e.g.,
instructions that are correct but different in prose will receive a
low BLEU score). In an effort to further evaluate the accuracy and
usability of our method, we conducted a set of human evaluation experiments 
in which we asked $42$ novice participants on Amazon Mechanical Turk 
($21$ females and $21$ males, ages
$18$--$64$, all native English speakers) to follow natural language
route instructions, randomly chosen from two equal-sized sets of
instructions generated by our method and by humans for $50$ distinct
paths of various lengths. The paths and corresponding human-generated
instructions were randomly sampled from the SAIL test set.  Given a
route instruction, human participants were asked to navigate to the
best of their ability using their keyboard within a first-person,
three-dimensional virtual world representative of the three
environments from the SAIL corpus. Fig.~\ref{fig:simulator} provides
an example of the participants' field of view while following route
instructions. After attempting to follow each instruction, each
participant was given a survey composed of eight questions, three
requesting demographic information and five requesting feedback on
their experience and the quality of the instructions that they
followed. We collected data for a total of $441$ experiments ($227$ 
using human annotated instructions and $214$ using machine generated 
instructions). 
The system randomly assigned the experiments
to discourage the participants from learning the environments or becoming
familiar with the style of a particular instructor. No participants
experienced the same scenario with both human annotated and machine
generated instructions. Appendix~\ref{app:human_subjects_eval} provides further 
details regarding the experimental procedure.

\section{Results} \label{sec:results}

We evaluate the performance of our architecture by scoring 
the generated instructions using the $4$-gram BLEU score commonly used
as an automatic evaluation mechanism for machine
translation. Comparing to the human-generated instructions, our
method achieves sentence- and corpus-level 
BLEU scores of $74.67\%$ and $60.10\%$, respectively, on the validation set. 
On the test set, the
method achieves sentence- and corpus level BLEU scores of $72.18\%$
and $45.39\%$, respectively. Fig.~\ref{fig:generation_example} shows an 
example of a route instruction  generated by our system for a given map and path.

\subsection{Aligner Ablation}
\begin{figure}[!t]
    \centering
    \includegraphics[width=\linewidth]{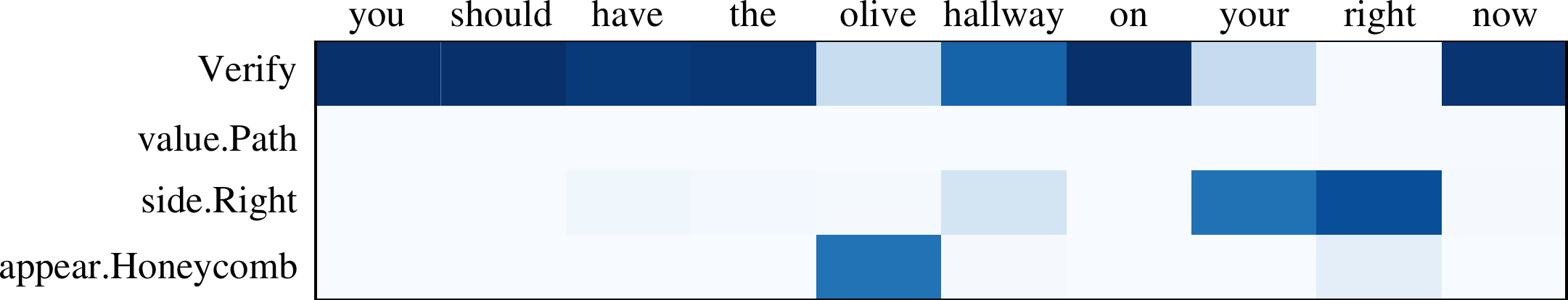}\\[6pt]
    \includegraphics[width=\linewidth]{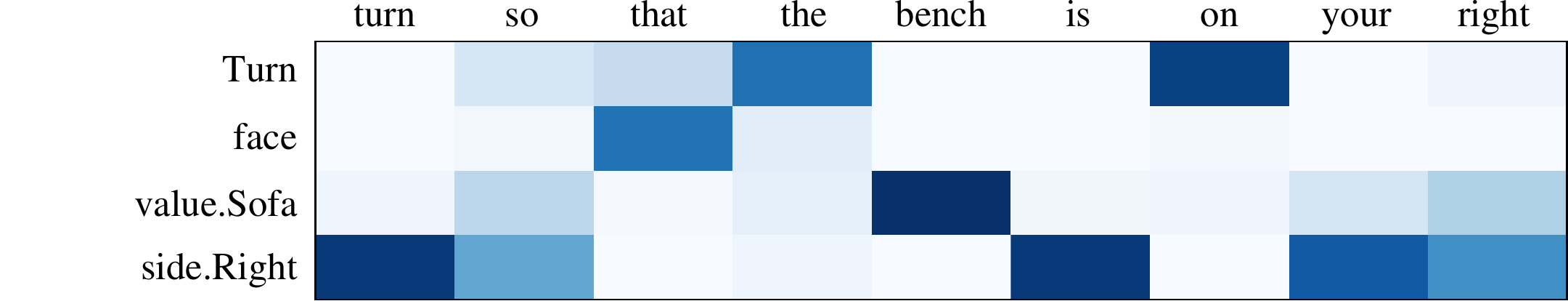}
    \caption{Alignment visualization for two
      pairs of CAS (left) and natural language instructions (top). Darker colors
      denote greater attention weights.
		\label{fig:attention_heatmap}}
\end{figure}
Our model employs an aligner in order to learn to focus on particular CAS tokens
that are salient to words in the output instruction. We evaluate the
contribution of the aligner by implementing and training an
alternative model in which the last encoder hidden state is fed
to the decoder. Table~\ref{tab:ablation_test} compares
the performance of the two models on the original validation set. The
inclusion of an aligner results in a slight increase in the BLEU score of the generated
instructions relative to the human-provided references, and is also
useful as a means of visualizing the inner workings of our model (as
shown below). Additionally, we empirically
find that the aligner improves our model's ability to learn the association
between CAS elements and words in the output, thereby yielding better instructions.
\begin{table}[h]
    \centering
    \begin{tabularx}{0.85\linewidth}{r c c}
      \toprule
      & Full Model & No Aligner\\
      \midrule
      sentence-level BLEU & $\mathbf{74.67}$  & $74.40$ \\
      corpus-level BLEU & $\mathbf{60.10}$  & $57.40$ \\
      \bottomrule
    \end{tabularx}
    \caption{Aligner ablation results.}\label{tab:ablation_test}
\end{table}

\subsection{Language Model Ablation}

Our method employs a language model to rank  instructions generated for the 
different candidate CAS commands and across different settings
of the beam width. In practice, the language model, trained on large amounts of 
English data, helps to remove grammatically incorrect sentences produced by the 
sequence-to-sequence model, which is only trained on the smaller pairwise dataset. 
Table~\ref{tab:lm_ablation_test} presents two instruction
candidates generated by our encoder-aligner-decoder model for two
different CAS commands. Our language model
successfully assigns high perplexity scores to the incorrect
instructions, with the chosen instruction being grammatically correct.
\begin{table}[h]
    \centering
    \begin{tabularx}{\linewidth}{r l}
      \toprule
      LM-score & Candidate
      \\
      \midrule
      $105.00$ & ``so so a straight chair to your left'' \\
      $27.65$ & ``turn so that the chair is on your left side'' \\
      \midrule
      $101.00$ & ``keep going till the blue flor id on your left'' \\
      $11.00$ & ``move until you see blue floor to your right'' \\
      \bottomrule
    \end{tabularx}
    \caption{Language model ablation outputs}\label{tab:lm_ablation_test}
\end{table}

\subsection{Aligner Visualization}

Figure~\ref{fig:attention_heatmap} presents heat maps that visualize the
alignment between a CAS command input into surface realization (left)
and the generated route instruction (top) for two different
scenarios drawn from the SAIL validation set. The visualizations
demonstrate that our method learns to align elements of the formal CAS
command with their corresponding words in the generated
instruction. For example, the network learns the association between
the honeycomb textured floor and its color (top); that ``bench'' refers to
sofa objects (bottom); and that the phrase ``you should have'' indicates a
verification action (top).

\subsection{Human Evaluation}
\begin{figure}[!t]
    \centering
    \includegraphics[width=\linewidth]{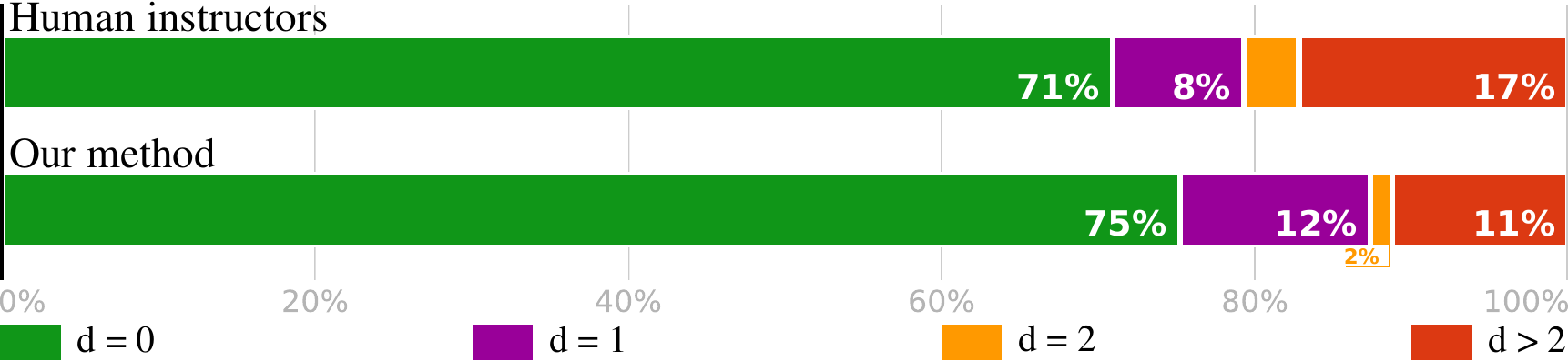}
    \caption{Comparison between the performances achieved by the participants 
    while following human annotated and machine generated instructions.
    \label{fig:study_performance}}
\end{figure}
We evaluate the accuracy with which human participants followed the
natural language
instructions in terms of the Manhattan
distance $d$ between the desired destination (i.e., the last pose of the 
target path) and the participant's location when s/he finished the scenario. 
Figure~\ref{fig:study_performance} compares the accuracy of the
participants' paths when following human-generated instructions (i.e.,
those from the SAIL test dataset) with those corresponding to
instructions that our method produced. We report the fraction of times
that participants finished within different distances from the
goal.\footnote{We note that the $d=0$ accuracy for the human-generated
  instructions is consistent with that reported
  elsewhere~\cite{chen11}.} The results demonstrate that participants
reached the desired position $4\%$ more often when following
instructions generated using our method compared against the human
instruction baseline. When they didn't reach the destination,
participants reached a location within one vertex away $8\%$ more
often given our instructions. Meanwhile our method yields a failure
rate ($d>2$) that is $6\%$ lower. Note that of scenarios in which
participants reached the destination, the total time required to
interpret and follow our method's instructions is
$9.52$\,s less than
that of the human-generated instructions, though the difference is not
statistically significant.

\begin{figure}[t!]
    \centering
	\subfigure[{\small Q1:~``{How do you define the amount of information provided?}''}]{
    	\label{fig:survey_information}
    	\includegraphics[width=\linewidth]{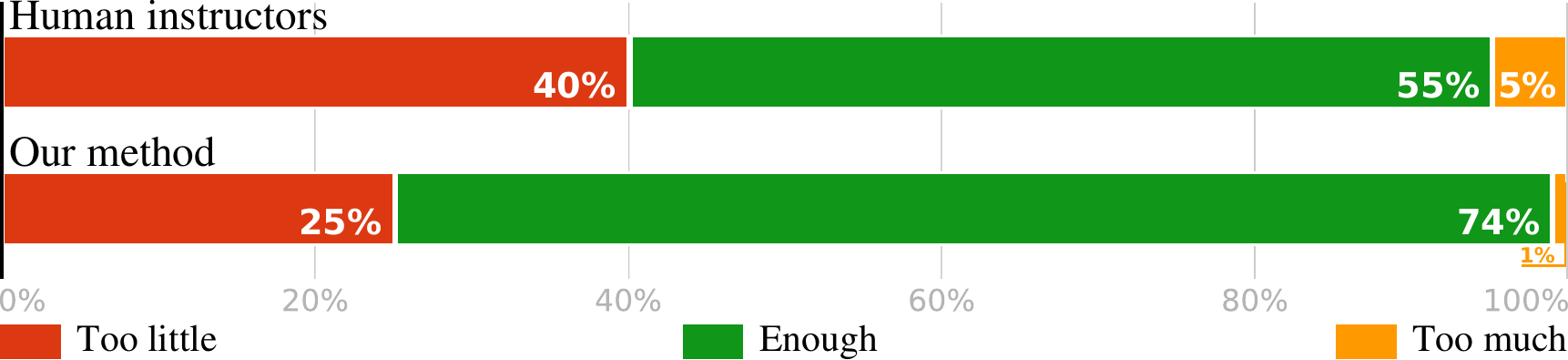}
    }\\[14pt]
    \subfigure[{\small Q2:~``{How would you evaluate the task in terms of difficulty?}''}]{
    	\label{fig:survey_difficulty}
    	\includegraphics[width=\linewidth]{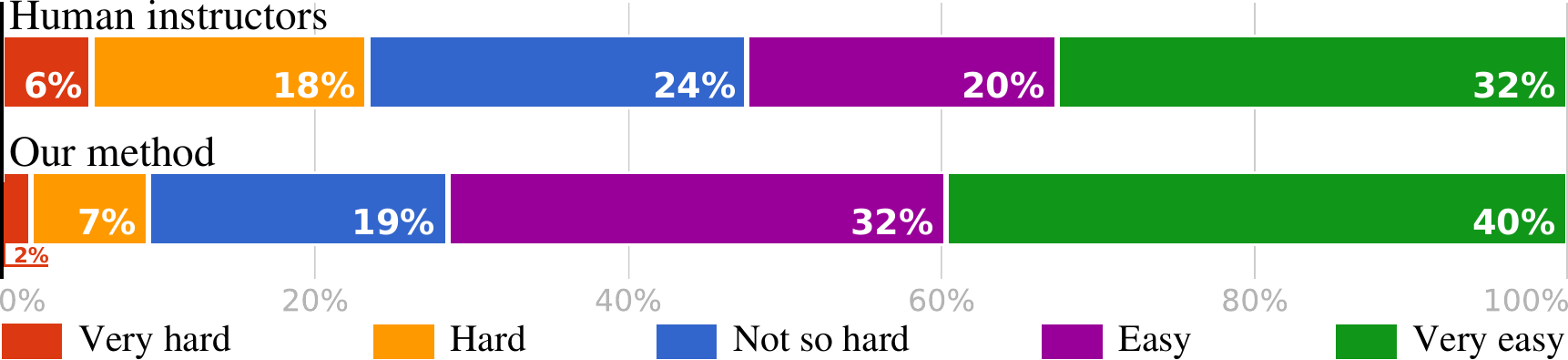}
    }\\[14pt]
    \subfigure[{\small Q3:~``{How confident are you that you followed the desired path?}''}]{
    	\label{fig:survey_confidence}
    	\includegraphics[width=\linewidth]{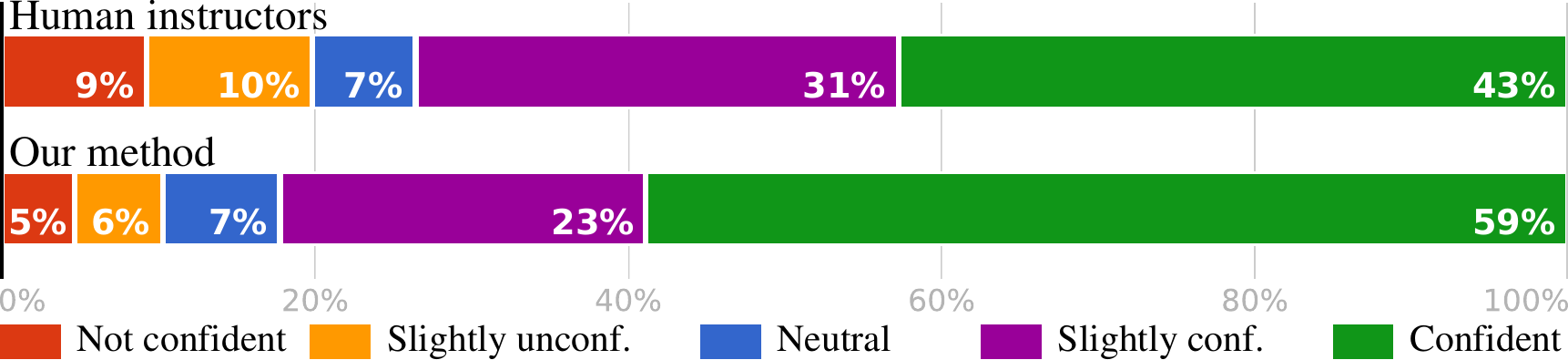}
    }\\[14pt]
    \subfigure[{\small Q4:~``{How many times did you have to backtrack?}''}]{
    	\label{fig:survey_backtrack}
    	\includegraphics[width=\linewidth]{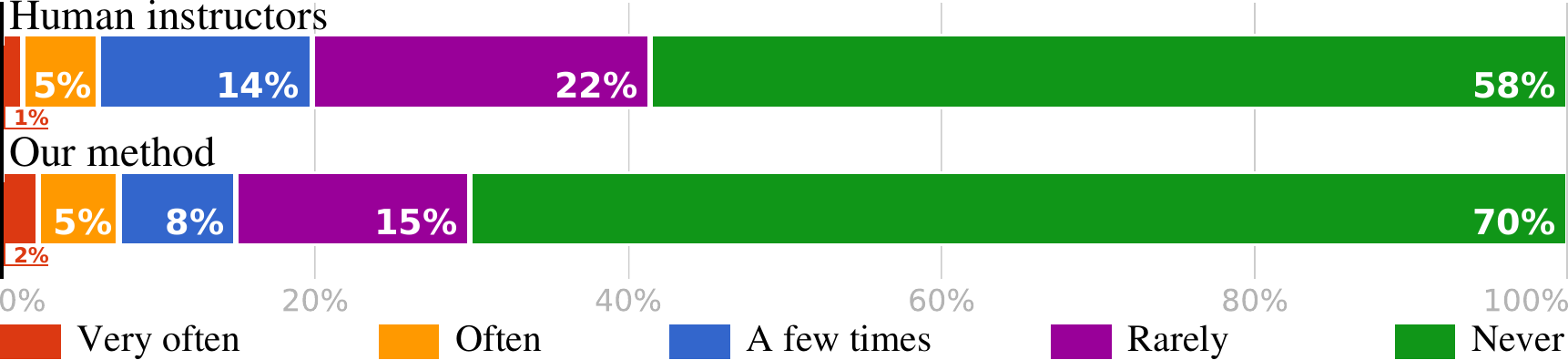}
    }\\[14pt]
    \subfigure[{\small Q5:~``{Who do you think generated the instructions?}''}]{
    	\label{fig:survey_who}
    	\includegraphics[width=\linewidth]{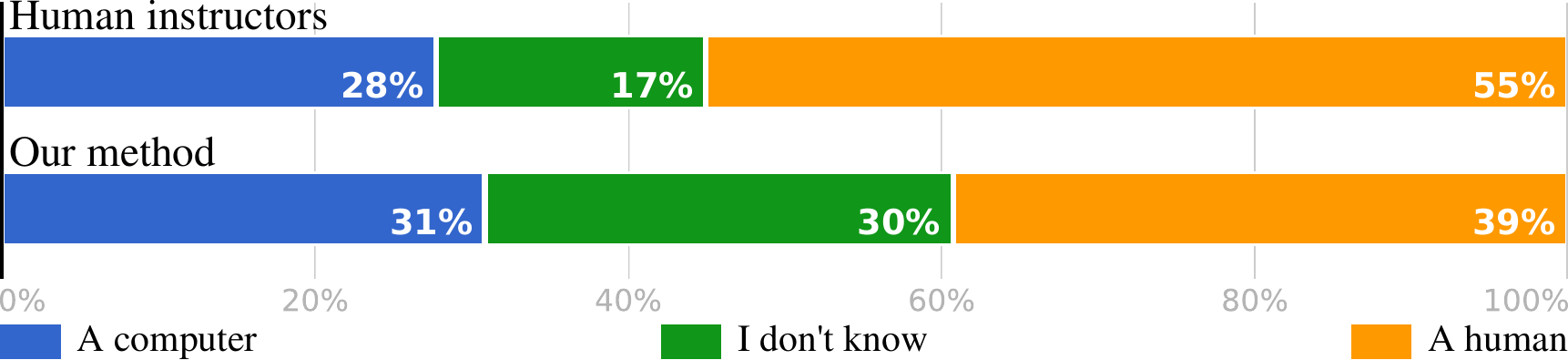}
    }
    \caption{Participants' survey response statistics.
		\label{fig:survey_results}
	}
\end{figure}
Figure~\ref{fig:survey_results} presents the participants' responses to
the survey questions that query their experience following the
instructions. By using IRL to learn a content selection policy for
constructing CAS structures, our method generates instructions that
convey enough information to follow the command and were rated as
providing too little information $15\%$ less frequently than the
human-generated baseline
(Fig.~\ref{fig:survey_information}). Meanwhile, participants felt that
our instructions were easier to follow
(Fig.~\ref{fig:survey_difficulty}) than the human-generated baselines
($72\%$ vs.\ $52\%$ rated as ``easy'' or ``very easy'' for our method
vs.\ the baseline).  Participants were more confident in their ability
to follow our method's instructions (Fig.~\ref{fig:survey_confidence})
and felt that they had to backtrack less often
(Fig.~\ref{fig:survey_backtrack}). Meanwhile, both types of
instructions were confused equally often as being machine-generated
(Fig.~\ref{fig:survey_who}), however participants were less sure of
who generated our instructions relative to the human baseline.

\begin{figure}[t]
\centering
\begin{tabularx}{\linewidth}{c c X}
	\toprule
	\multicolumn{3}{l}{\textbf{Map and Paths}}\\
	\multicolumn{3}{c}{%
	\centering
	\includegraphics[width=0.96\linewidth]{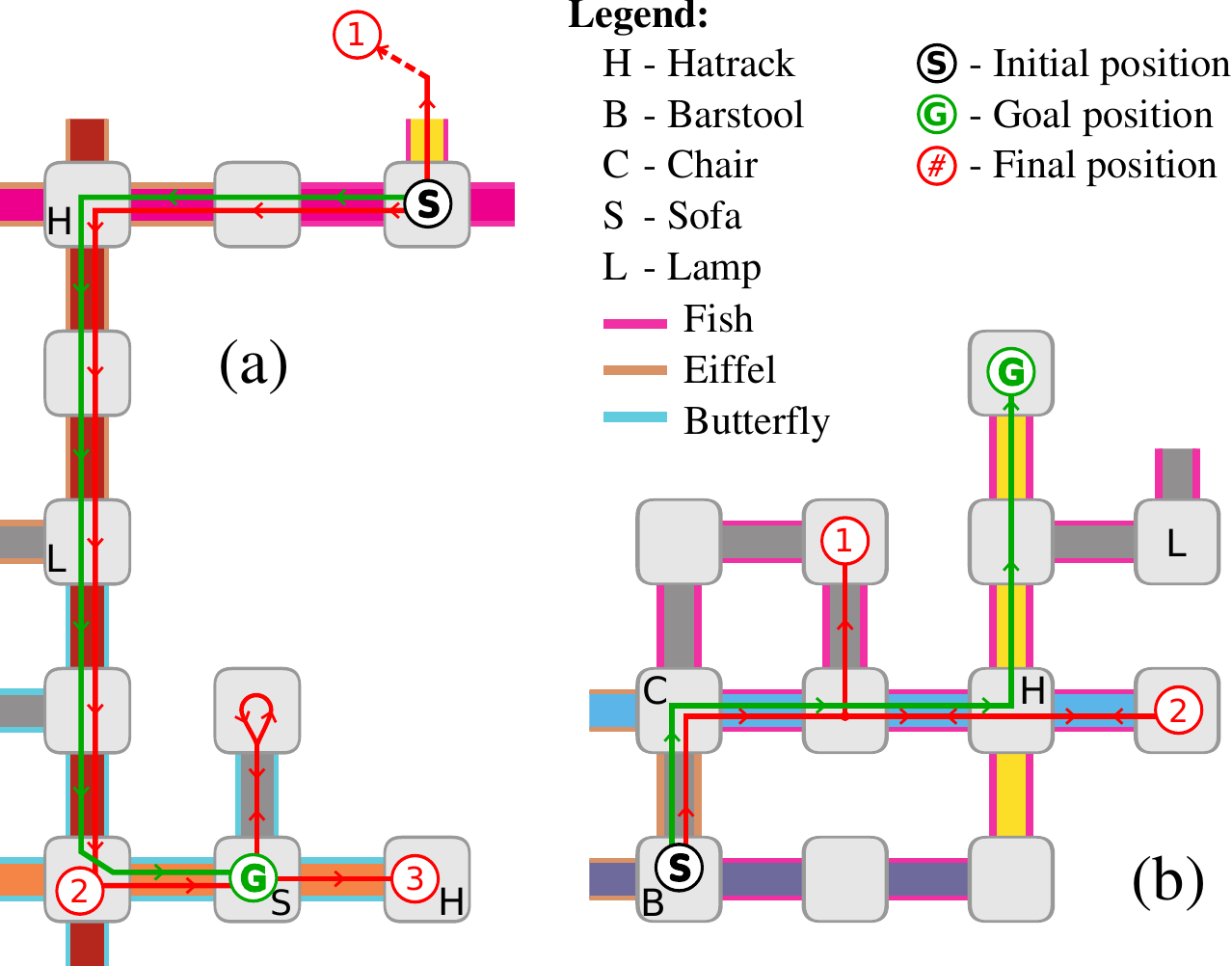}
	}\\
	\midrule
	\multicolumn{3}{l}{\textbf{Instructions}}\\
	\multirow{2}{*}[-2.5em]{ {\large (a)} } 
	& Human & ``with your back to the wall turn left. walk along the
              flowers to the hatrack. turn left. walk along the brick
              two alleys past the lamp. turn left. move along the
              wooden floor to the chair. in the next block is a
              hatrack''\\[3pt]%
  \cline{2-3}\\[-5pt]
	& Ours & ``you should have the olive hallway on your right now. walk forward twice. turn left. move until you see wooden floor to your left. face the bench. move to the bench''\\
	\midrule
	\multirow{2}{*}[-2.0em]{ {\large (b)} } 
	& Human & ``head toward the blue floored hallway. make a right on
              it. go down till you see the fish walled areas. make a
              left in the fish walled hallway and go to the very
              end''\\[3pt]
  \cline{2-3}\\[-5pt]
	& Ours & ``turn to face the white hallway. walk forward once. turn right. walk forward twice. turn left. move to the wall''\\
    \bottomrule
\end{tabularx}
\caption{Examples of paths from the SAIL corpus that ten participants
  (five for each map) followed according to
  instructions generated by humans and by our method. Paths in red are
those traversed according to human-generated instructions, while paths
in green were executed according to our instructions. Circles with
an ``S'' and ``G'' denote the start and  goal locations, respectively.}\label{fig:generation_examples}
\end{figure}
Figure~\ref{fig:generation_examples} compares the paths that
participants took when following our instructions with those that they
took given the reference human-generated directions. In the case of
the map on the left (Fig.~\ref{fig:generation_examples}(a)), none of the five participants reached the correct
destination (indicated by a ``G'') when following the human-generated
instruction. One participant reached location $2$, three
participants stopped at location $3$ (one of whom backtracked after
reaching the end of the hallway above the goal), and one participant
went in the wrong direction at the outset. In 
contrast, all five participants reached the goal directly (i.e.,
without backtracking) when following our instruction. For the scenario depicted on the right (Fig.~\ref{fig:generation_examples}(b)), five
participants failed to reach the destination when provided with the
human-generated instruction. Two of the participants went  directly to
location $1$, two participants navigated to location $2$, and one participant went to
location $2$ before backtracking and taking a right to location
$1$. We attribute the failures to the ambiguity in the human-generated
instruction that references ``fish walled areas,'' which could
correspond to most of the hallways in this portion of the map (as
denoted by the pink colored lines). On the other hand, each of the
five participants followed the intended path (shown in green) and
reached the goal when following the instruction generated using our
method. We note that the second-best candidate that our framework
considered mentions the ``olive hallway'' as a unique reference to
where the person should take a left.
\section{Conclusion} 
\label{sec:conclusion}

We presented a model for natural language generation in the context of
providing indoor route instructions that exploits a structured
approach to produce unambiguous, easy to remember and grammatically
correct human-like route instructions. Currently, our model generates
natural language route instructions for the shortest path to the
goal. Nevertheless, there are situations in which a longer path may
afford instructions that are more
straightforward~\cite{richter2008simplest} or that increase the
likelihood of reaching the
destination~\cite{haque2006algorithms}. Another interesting direction
for a future work would be to involve the integration of a model of
instruction followers~\cite{mei2016listen} with our
architecture in an effort to learn to generate instructions that are
easier to follow. Such an approach would permit training the model in
a reinforcement learning setting, directly optimizing over task performance.
\section{Acknowledgements}

We gratefully acknowledge the support of the NVIDIA Corporation for the
donation of the Titan X GPU used for this research.


\begin{appendices}
\section{MDP Feature Representation}
\label{app:mdp_feature}
\begin{table}[!hp]
    \centering
    \begin{tabularx}{0.85\linewidth}{l  l}
      \toprule
      \textbf{Context} & \textbf{Description} (binary)\\
      \midrule
      {\tt t} & change orientation \\
      {\tt w} & change position \\
      {\tt tw} & change orientation and then position \\
      {\tt wt} & change position and then orientation \\
      {\tt w\_obj\_at} & the final place contains an object \\
      {\tt w\_past\_obj} & pass an object while walking \\
      {\tt w\_dead\_end} & the final place is a dead-end \\
      {\tt w\_goal} & the final pose is the goal pose \\
      {\tt t\_start} & it is the first action to take \\
      {\tt t\_new\_carp} & final pose faces a new floor color \\
      {\tt t\_obj\_side} & an object is visible from the final pose \\
      {\tt t\_obj\_at} & there is an object at the turn location \\
      {\tt t\_new\_pict} & final pose faces a new wall color \\
      {\tt t\_at\_T} & the place where to turn at is a dead-end \\
      \bottomrule
    \end{tabularx}
  \caption{Contexts used as path features\label{tab:path_contexts}}
\end{table}
\begin{table}[!h]
    \centering
    \begin{tabularx}{0.825\linewidth}{c l}
        \toprule
        \textbf{Property} & \textbf{Description} \\
        \midrule
		{\tt nsl} & number of key information to remember \\
        {\tt cmd} & low-level command groundtruth \\
        {\tt dep} & CAS command maximum depth \\
        {\tt eta} & number of defined attributes \\
        {\tt pcp} & number of floor colors mentioned \\
        {\tt ppc} & number of wall colors mentioned \\
        {\tt htw} & whether or not to head towards an object \\
        {\tt nln} & number of landmarks mentioned \\
		{\tt trf} & turn reference frame \\
        \bottomrule
    \end{tabularx}
    \caption{Properties used as CAS structure features\label{tab:cas_properties}}
\end{table}
We use $14$ \textit{contexts} as features for paths and $9$ instruction
\textit{properties} as features for CAS structures. For each
demonstration, \textit{map} and \textit{path} are represented by a
single binary vector of 14 elements (indicating which contexts are
active and which are not) while the \textit{instruction} is
represented by an integer-valued vector of 9 elements. The lists of
contexts and instruction properties we use in our model are shown in
Table~\ref{tab:path_contexts} and \ref{tab:cas_properties}
respectively.
\section{Human Subjects Evaluation}
\label{app:human_subjects_eval}

We evaluated the accuracy and quality of our generated instructions
via a set of experiments in which human participants were asked to
navigate a three-dimensional virtual environment according route
instruction that was provided.  Participants were recruited using the
Amazon Mechanical Turk crowd-sourcing platform. The recruiting message
said that the objective of the experiment was to understand how people
follow route instructions.  We offered
\$$0.15$~(USD) for each completed scenario.  Fifty-four people
participated and completed a total of
$511$ experiments.  We omitted those experiments for
which the participant answered ``No'' or ``Do not disclose'' to the
question ``Are you a native English speaker?'', since it was included
as a requirement for participating. This procedure resulted in a total of
$42$ participants ($21$ females and $21$ males, ages
$18$--$64$) and a total of
$441$ experiments.  We did not omit experiments based
on the participants' performance or the answers they gave to questions
regarding demographic information.  We paid all the participants for
their contribution, regardless of whether they were native English
speakers. Prior to taking part, each
participant spent at least $30$ seconds navigating within a held-out
environment in order to familiarize themselves with the
interface. Each experimented lasted an average of $40$ seconds. The
route instructions were randomly sampled from 
those generated using our method and those provided by humans as part
of the SAIL corpus.  The following outlines the procedure that each
participant then followed:
\begin{enumerate}
\item The participant was presented with a virtual environment in
    which s/he was placed at the start position facing a random
    orientation, and given the route instructions.
\item The participant was asked to navigate according
    to the instruction using their keyboard's arrow keys.
\item At any time, the participant could review the directions
    and a legend containing information about objects, pictures and
    floor colors found in the environment.
\item When the participant believed that s/he had reached the
    destination, s/he pressed the ``Finish task'' button.
\item The participant was presented with a survey consisting 
    of eight questions, three requesting demographic information and
    five requesting feedback on their experience and the quality of
    the instructions that they followed.
\end{enumerate}
\end{appendices}

\bibliographystyle{abbrvnat}
{\small%
	\bibliography{references}%
}

\end{document}